\documentclass[twoside,11pt, preprint]{article} 
\usepackage{jmlr2e_preprint}

\usepackage{lastpage}
\usepackage{subcaption, caption, adjustbox}
\usepackage{tabularx}
\usepackage{bm}
\usepackage{bbm}
\usepackage{mathtools}
\usepackage{wrapfig}
\usepackage{xurl}

\usepackage{listings, color, wrapfig}
\usepackage{xcolor}

\definecolor{dkgreen}{rgb}{0,0.6,0}
\definecolor{gray}{rgb}{0.5,0.5,0.5}
\definecolor{mauve}{rgb}{0.58,0,0.82}

\newcommand{\delpv}[1]{{\color{gray}{}}}

\newcommand{\revision}[1]{{\color{black} #1}}

\jmlrheading{X}{2023}{1-X}{XX}{XX/23}{00-000}{}

\ShortHeadings{\texttt{ODTlearn}: A Package for Learning Optimal Decision Trees}{Vossler et al}
\firstpageno{1}

\usepackage{standalone}

\begin{document}

\title{\texttt{ODTlearn}: A Package for Learning Optimal Decision Trees for Prediction and Prescription}

\author{\name Patrick Vossler\thanks{Both authors contributed equally.} \email pvossler@usc.edu \\
       \name Nathan Justin\footnotemark[\value{footnote}] \email njustin@usc.edu \\
       \name Sina Aghaei \email saghaei@usc.edu\\
       \name Nathanael Jo \email nathanael.jo@gmail.com\\
       \name Andrés Gómez \email gomezand@usc.edu\\
       \name Phebe Vayanos \email phebe.vayanos@usc.edu\\
       \addr University of Southern California, Center for AI in Society, Los Angeles, CA 90089
       }

\editor{}

\maketitle

\begin{abstract}

\texttt{ODTlearn} is an open source Python package that provides methods for learning optimal decision trees for high-stakes predictive and prescriptive tasks based on the state-of-the-art mixed-integer optimization (MIO) framework proposed \revision{by \cite{aghaei2025strong}}. The current version of the package provides implementations for learning optimal classification trees, optimal fair classification trees, optimal prescriptive trees from observational data, and optimal classification trees robust to distribution shifts. We have designed the package to be easy to maintain and extend as new optimal decision tree problem classes, reformulation strategies, and solution algorithms are introduced. To this end, the package follows object-oriented design principles and supports both commercial (Gurobi) and open source (COIN-OR branch and cut) solvers. The package documentation, user guide, installation instructions, link to source code, and instructions for submitting bug reports and feature requests can all be found at \url{https://d3m-research-group.github.io/odtlearn/}. 
\end{abstract}


\section{Introduction}
\label{sec:intro}

Automated data-driven predictive and prescriptive methods are increasingly being used to inform and support decision-making in high-stakes domains.
In such settings, these tools should be:
(a) \textit{accurate} (to minimize erroneous predictions/prescriptions that may harm the populations on which they are deployed), (b) \textit{interpretable} (so that predictions and decisions are transparent, accountable, and easy to audit), (c) \textit{flexible} (i.e., easily augmented with domain specific constraints such as capacity and/or fairness constraints), and (d) \textit{robust} (to ensure high-quality solutions even under adversarial shifts between training and deployment data) \citep{ toreini2020relationship,Varshney2022, justin2025responsible}.


Decision trees are a popular supervised learning model used for both predictive and prescriptive tasks and typically taking the form of a binary tree. Each branching node tests one feature and directs a datapoint left or right based on the outcome, while each leaf assigns a predicted label or a treatment. Especially for shallow trees, this structure makes them easy to visualize, use, and scrutinize, and therefore highly interpretable. While standard, non-MIO-based decision tree learning algorithms such as CART~\citep{breiman1984classification}, ID3~\citep{quinlan1986induction}, C4.5~\citep{quinlan2014c4}, and Policy Tree~\citep{zhou2023offline} are extremely fast, they may produce suboptimal trees with inferior test set performance and/or cannot directly accommodate domain-specific considerations such as e.g., fairness or capacity constraints, or different kinds of objectives, such as worst-case accuracy under shift. Thus, they necessitate dedicated heuristics/algorithms to be designed before they can be applied to these settings. Over the last decade, motivated by these limitations and by the tremendous improvements in MIO technology, MIO-based approaches to learning optimal trees have burgeoned~\citep{bertsimas2017optimal,verwer2019learning,aghaei2019learning,aghaei2025strong,jo2026learning}. 
\revision{While often significantly slower to train, even for shallow trees, their optimality guarantees and modeling flexibility make many of these MIO-based methods attractive and preferred for a variety of applications.}

\revision{The \texttt{ODTlearn} Python package provides methods for fitting provably optimal decision trees for prediction and prescription using highly versatile, fast, and adaptable MIO technology. It is based on the ``flow-based'' FlowOCT formulation of \citet{aghaei2025strong} for learning optimal classification trees and its variants, all of which are widely considered as the state-of-the-art MIO approaches for learning optimal trees. The defining feature of these approaches is that they convert the decision tree to a capacitated flow graph, where the path of each data sample through the tree can be identified by solving a maximum flow problem. This idea, originally proposed by~\citet{aghaei2025strong}, decouples sample traversal from the optimization objective, making flow-based formulations highly versatile, being easy to augment with constraints on individual samples or groups of samples and to optimize alternative objectives. It also results in a provably stronger MIO formulation than previous MIO methods and allows FlowOCT (and several of its variants) to have a structure amenable to decomposition techniques, such as BendersOCT, a tailored Benders' decomposition approach. As a result, FlowOCT and BendersOCT outperform all prior MIO-based formulations for learning optimal classification trees in terms of both accuracy and speed. Across datasets from the UCI Machine Learning Repository~\citep{Dua2017UCIRepository}, \citet{aghaei2025strong} report that BendersOCT is, on average, up to 8\% and 36\% more accurate than BinOCT \citep{verwer2019learning} and OCT \citep{bertsimas2017optimal}, respectively, while being 29 times faster.}

\revision{The versatility of flow-based formulations has allowed them to be used for learning optimal fair classification trees (FairOCT, \citet{jo2023learning}), optimal trees robust to distribution shifts (RobustOCT, \citet{justin2021optimal, justin2023learning}), and optimal prescriptive trees (FlowOPT, \citet{jo2026learning}), all implemented in \texttt{ODTlearn}.}
FairOCT adds group fairness constraints such as statistical parity, conditional statistical parity, or equalized odds to FlowOCT. On widely-studied fairness datasets in the literature~\citep{angwin2016machine, Dua2017UCIRepository}, FairOCT achieves near-perfect fairness while only losing on average 4.2 percentage points in out-of-sample accuracy relative to the best-performing complex machine learning models~\citep{jo2023learning}. RobustOCT augments FlowOCT with a robust objective, leveraging the flow formulation to model how distribution shifts may alter the paths taken by data samples through the tree. RobustOCT improves accuracy by up to 12.48 percentage points over BendersOCT under distribution shifts between the training and testing data \citep{justin2023learning}. FlowOPT adapts FlowOCT to learning prescriptive trees that assign treatments to maximize a statistically consistent estimate of expected outcomes. 
On simulated observational treatment assignment settings based on warfarin data \citep{international2009estimation}, \cite{jo2026learning} show that FlowOPT achieves an average out-of-sample probability of correct treatment assignment that is 29 and 32 percentage points higher than the MIO-based prescriptive tree formulations by~\cite{kallus2017recursive} and~\cite{bertsimas2019optimal}, respectively. At the same, FlowOPT can conveniently incorporate domain-specific constraints such as capacity constraints or even allocation or outcome fairness constraints. 

The remainder of this paper is organized as follows. Section~\ref{sec:comparison} compares \texttt{ODTlearn} to related software. Section~\ref{sec:implementation} describes its architecture, usage, and development practices. 

\section{Comparison to Related Software}
\label{sec:comparison}
A wide range of software packages exists for decision tree learning, differing in their optimality guarantees, applicability, extensibility, and accessibility. In the following, we review the software most closely related to \texttt{ODTlearn}.

\revision{
\textbf{Optimal Decision Tree Codebases.} There are several open source implementations for learning optimal decision trees using dynamic programming-based \citep{JMLR:v23:20-520} and MIO-based approaches, with the latter being most closely related to \texttt{ODTlearn}. Although many of these are not distributed as a package and created primarily for reproducibility purposes, they provide valuable implementations of optimal methods for technical users and researchers.}
\revision{The authors of the BinOCT and FlowOCT MIO formulations provide open-source implementations of their methods, see \citet{verwer_binoct} and \citet{aghaei_strongtree}, respectively. \cite{BoTangOCT} provide open-source implementations of the BinOCT, OCT, and FlowOCT formulations. However, none of these implementations is distributed as a Python package or indexed on PyPI, limiting their accessibility and ease of adoption. Moreover, the implementations primarily rely on proprietary MIO solvers, such as CPLEX and Gurobi, which require paid or academic licenses to solve realistic-scale problems. Finally, their codebases are designed around specific formulations, limiting extensibility and adaptability to new problem settings. \texttt{ODTlearn} in contrast builds upon the highly-extensible flow-based formulations (see section \ref{sec:intro}), uses dedicated object-oriented programming principles for developing software that is easy to maintain and extend, and supports both open-source and commercial solvers (see section \ref{sec:arch}).} 

\textbf{Packages for Learning Decision Trees.}
Some alternative packages for learning decision trees in R and Python include
\texttt{scikit-learn}~\citep{pedregosa2011scikit}, \texttt{rpart}~\citep{therneau1997introduction}, \texttt{caret}~\citep{kuhn2008caret}, and
\revision{\texttt{evtree}~\citep{Grubinger2014-zl}},
among others. These packages rely on greedy algorithms such as CART, ID3, and C4.5 or customized heuristic methods which, as discussed in section~\ref{sec:intro}, lack global optimality guarantees or provide limited modeling flexibility.
\revision{Interpretable AI~\citep{InterpretableAI} offers a proprietary Julia~\citep{Julia-2017} package with a Python wrapper that provides methods for learning classification, prescriptive, and regression trees. It incorporates local search and random restarts to improve solution quality, without providing a certificate of optimality.}
There are also some open-source Python packages that implement non-MIO approaches for learning optimal decision trees.
\revision{The \texttt{PyDL8.5}~\citep{aglin2021pydl8} package implements the DL8.5~\citep{aglin2020learning} itemset mining approach, and \texttt{gosdt}~\citep{Lin2020-hv, mctavish2022fast} provides a C++ implementation of Generalized Optimal Sparse Decision Trees using branch-and-bound with dynamic programming.}
The \texttt{policytree} package~\citep{Sverdrup2020-jk} provides an R implementation of a global tree-search approach by \citet{zhou2023offline}, which learns optimal prescriptive trees from observational data.
None of these packages support constructing trees subject to domain-specific constraints, while heuristic-based packages do not provide optimality gaps for the solutions they produce.

\section{Software Design, Usage, and Development}
\label{sec:implementation}
The applicability, flexibility, accessibility, and extensibility of \texttt{ODTlearn} are enabled by its software architecture, intuitive usage design, and open-source development model.

\subsection{Software Architecture} \label{sec:arch}
\begin{figure}[ht]
    \centering
     \includegraphics[width=0.82\textwidth]{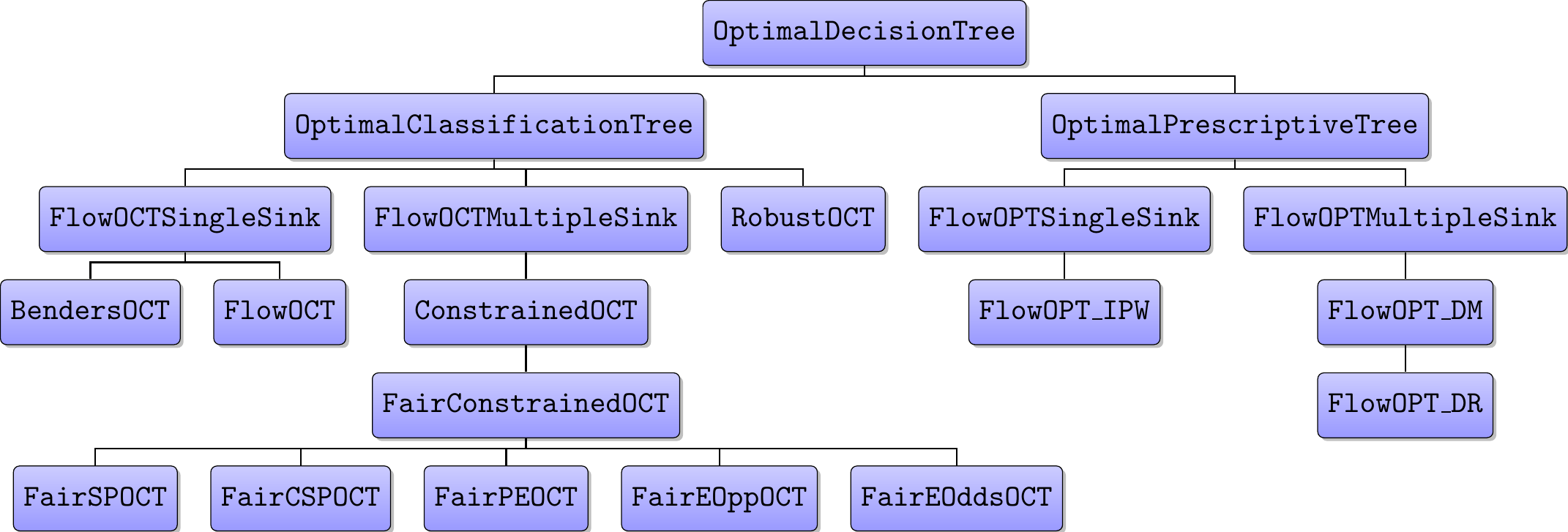}
\caption{Inheritance diagram for \texttt{ODTlearn}. \texttt{FlowOCT}/\texttt{BendersOCT}, \texttt{RobustOCT}, and the subclasses of \texttt{FairConstrainedOCT} implement the formulations of \citet{aghaei2025strong}, \citet{justin2021optimal, justin2023learning}, and \citet{jo2023learning}, respectively. \texttt{FlowOPT\_IPW}, \texttt{FlowOPT\_DM}, and \texttt{FlowOPT\_DR} implement the prescriptive tree methods of \citet{jo2026learning}.}
    \label{fig:simple_diagram}
\end{figure}

The software architecture of \texttt{ODTlearn} is based on the class structure \revision{shown in Figure~\ref{fig:simple_diagram} and} follows the SOLID principles of object-oriented programming for developing maintainable and extensible software~\citep{martin2003agile}.
\revision{The Single Responsibility Principle (SRP) states that each class should have a single purpose. The Open-Closed Principle (OCP) requires that a class should be extensible without modifying its existing implementation. According to the Liskov Substitution Principle (LSP), derived classes should be able to replace their base classes without affecting correctness. The Interface Segregation Principle (ISP) asserts that interfaces should not be forced to depend on methods they do not use. Finally, the Dependency Inversion Principle (DIP) suggests that software design should rely upon abstract interfaces rather than concrete implementations.}

\revision{Following DIP, all classes in \texttt{ODTlearn} are derived from our abstract base class \texttt{Optimal\allowbreak DecisionTree}, which provides a standardized interface for the two types of trees currently supported (classification and prescription) while keeping their implementation details separate to follow ISP.
The \texttt{OptimalDecisionTree} abstract class allows our class structure to follow \revision{OCP} as we can create the derived classes \texttt{OptimalClassificationTree} and \texttt{OptimalPrescriptiveTree} with problem-specific methods for traversing and visualizing the decision tree.
The children of \texttt{OptimalClassificationTree} and \texttt{OptimalPrescriptive\allowbreak Tree} implement methods for creating the decision variables, the constraints, and the objective function necessary for constructing the optimization problem of interest.
The separate classes for each of the MIO formulations ensure that each of the derived classes has a single job, obeying SRP. Finally, the classes in the lower levels of Figure~\ref{fig:simple_diagram} implement user-facing methods such as \texttt{fit} and \texttt{predict} required by their parent classes. This ensures that our low-level classes respect LSP by adhering to the basic requirements of their parent class. Our adherence to the SOLID principles ensures that researchers and practitioners
can easily augment \texttt{ODTlearn} with more features (e.g., to learn trees optimizing different objectives or satisfying different constraints).}

\revision{Also aligned with the SOLID principles, \texttt{ODTlearn} includes an abstract base class, \texttt{Solver}, that provides a common solver interface and shared functionality for implementing MIO formulations.}
Currently, our package implements \texttt{GurobiSolver} and \texttt{CBCSolver}, both derived from the \texttt{Solver} class. These classes leverage the commercial Gurobi solver through \texttt{gurobipy} and the open-source Cbc solver~\citep{john_forrest_2026_18971618} through Python MIP~\citep{santos2020mixed}.
\revision{\texttt{ODTlearn} is designed such that the computational effort primarily lies within the chosen solver, as model construction and callback functions typically execute within a few seconds.}

\subsection{Usage}
\label{sec:usage}
\begin{figure}[ht]
    \centering
    \begin{minipage}[c]{0.5\textwidth}
    \begin{lstlisting}
    from odtlearn.flow_oct import BendersOCT
    import matplotlib.pyplot as plt

    clf = BendersOCT(depth=2, solver="gurobi",          time_limit=200)
    clf.fit(X, y) 
    
    clf.plot_tree()
    plt.show()
    \end{lstlisting}
    \end{minipage} \hspace{10mm}
    \begin{minipage}[c]{0.335\textwidth}         \includegraphics[width=\linewidth]{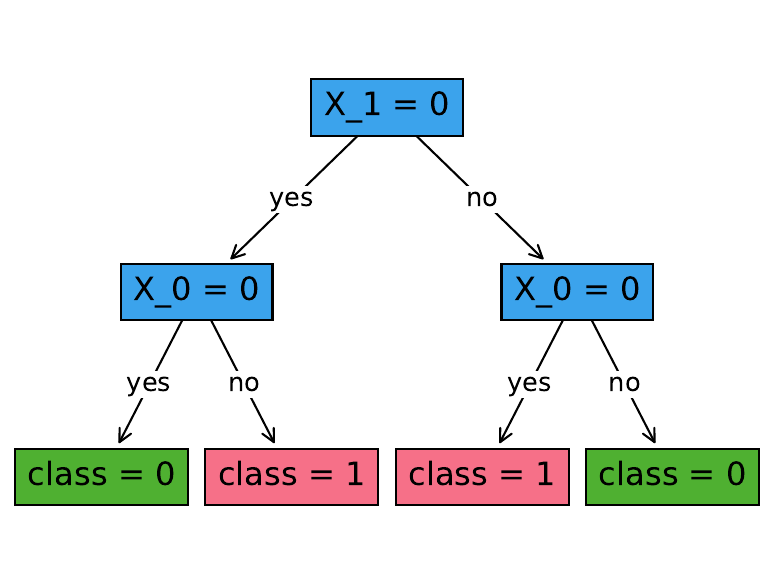}
    \end{minipage}
    \caption{\revision{Code for fitting an optimal classification tree on data \texttt{X} with labels \texttt{y}. The \texttt{BendersOCT} object specifies the solver and model parameters: \texttt{solver} indicates the solver used, \texttt{depth} restricts the depth of tree learned, and \texttt{time\_limit} sets the solver time limit in seconds. The \texttt{plot\_tree} function builds the figure on the right based on the learned tree.}}
    \label{fig:fair_example}
\end{figure}

To fit optimal decision trees, we provide users with a set of methods that follow a common \texttt{fit}/\texttt{predict} structure.
Figure~\ref{fig:fair_example} provides a code snippet showing how to use the \texttt{ODTlearn} API to fit \revision{an optimal classification tree using \texttt{BendersOCT}}. Once an optimal decision tree has been learned, users can employ the built-in \texttt{plot\_tree} function to visualize the tree.

\subsection{Development}
\label{sec:dev}
Releases of the \texttt{ODTlearn} package are available via PyPI at \url{https://pypi.org/project/ODTlearn}. The package source code and documentation are hosted on GitHub (\url{https://github.com/D3M-Research-Group/ODTlearn}). 
Our documentation (\url{https://d3m-research-group.github.io/odtlearn}) includes installation instructions, a user guide, an API reference, information on submitting bug reports and feature request, and example notebooks. The package is distributed under the GPL-3.0 license and makes use of several core libraries within Python’s scientific computing ecosystem: \texttt{scikit-learn}~\citep{pedregosa2011scikit}, \texttt{numpy}~\citep{harris2020array}, and \texttt{pandas}~\citep{mckinney2010data}.

\acks{P.\ Vayanos and N.\ Justin are funded in part by the National Science Foundation (NSF) under CAREER award
2046230. P.\ Vayanos and P.\ Vossler acknowledge support from the USC Zumberge Special Solicitation – Epidemic \& Virus Related Research and Development award.
N.\ Jo acknowledges support from the Epstein Institute at the University of Southern California. A.\ G\'omez is funded in part by NSF grant 2152777.
N.\ Justin is funded in part by the NSF Graduate Research Fellowship Program.
P. Vayanos and A. G\'omez are funded in part by the NSF NRT grant 2346058. They are grateful for the support.
}

\newpage
\bibliography{references}

\end{document}